\documentclass[journal]{IEEEtran}
\usepackage{amsmath}
\usepackage{amsthm}
\usepackage{amssymb}
\usepackage{algorithm}
\usepackage{array}
\usepackage{algpseudocode}
\usepackage{booktabs}
\usepackage{caption}
\usepackage{cite}
\usepackage{enumitem}

\usepackage{graphicx}
\usepackage{hyperref} 
\usepackage{multirow}
\usepackage{rotating} 
\usepackage{ragged2e} 
\usepackage{stfloats}
\usepackage{soul}
\usepackage{subcaption}
\usepackage{tabularx}
\usepackage{textcomp}
\usepackage{times}
\usepackage{url}
\usepackage{verbatim}
\usepackage{xcolor}

\newtheorem{definition}{Definition}

\newlist{tabitemize}{itemize}{1} 
\setlist[tabitemize]{label=\textbullet, nosep, leftmargin=*, 
                     before={\begin{minipage}[t]{\linewidth}\RaggedRight},
						after={\end{minipage}}}

\begin{document}

\title{LMEraser: Large Model Unlearning through Adaptive Prompt Tuning}

\author{Jie Xu, Zihan Wu, Cong Wang~\IEEEmembership{Fellow,~IEEE,} and Xiaohua Jia~\IEEEmembership{Fellow,~IEEE}
}

\markboth{Journal of \LaTeX\ Class Files,~Vol.~14, No.~8, August~2021}%
{Shell \MakeLowercase{\textit{et al.}}: A Sample Article Using IEEEtran.cls for IEEE Journals}


\maketitle

\begin{abstract}
	To address the growing demand for privacy protection in machine learning, we propose a novel and efficient machine unlearning approach for \textbf{L}arge \textbf{M}odels, called \textbf{LM}Eraser. Existing unlearning research suffers from entangled training data and complex model architectures, incurring extremely high computational costs for large models. LMEraser takes a divide-and-conquer strategy with a prompt tuning architecture to isolate data influence. The training dataset is partitioned into public and private datasets. Public data are used to train the backbone of the model. Private data are adaptively clustered based on their diversity, and each cluster is used to optimize a prompt separately. This adaptive prompt tuning mechanism reduces unlearning costs and maintains model performance. Experiments demonstrate that LMEraser achieves a $100$-fold reduction in unlearning costs without compromising accuracy compared to prior work. Our code is available at: \url{https://github.com/lmeraser/lmeraser}.
\end{abstract}


\section{Introduction} \label{sec:intro}
Large models such as BERT~\cite{kenton2019bert}, GPT-3~\cite{brown2020language}, and T5~\cite{raffel2020exploring} are characterized by their billions of parameters, complex architectures, and massive training data~\cite{lester2021power}. These characteristics enable them to recognize and learn complex patterns within data, achieving high accuracy and broad applicability. 

However, the increasing use of these models raises privacy concerns, as they may expose sensitive user information without permissions~\cite{carlini2022privacy,liu2022right}. Data privacy laws, such as the General Data Protection Regulation (GDPR) ~\cite{Voigt2017} and the California Consumer Privacy Act (CCPA) ~\cite{CCPA2023} grant users the `right to be forgotten,' permitting them to request removal of their personal data. \textit{Machine unlearning} has become a key solution to enforce this right, as it enables removing specific data from trained models without full retraining from scratch~\cite{xu2023machine}. 
\begin{figure}
	\centering
	\includegraphics[width=0.99\linewidth]{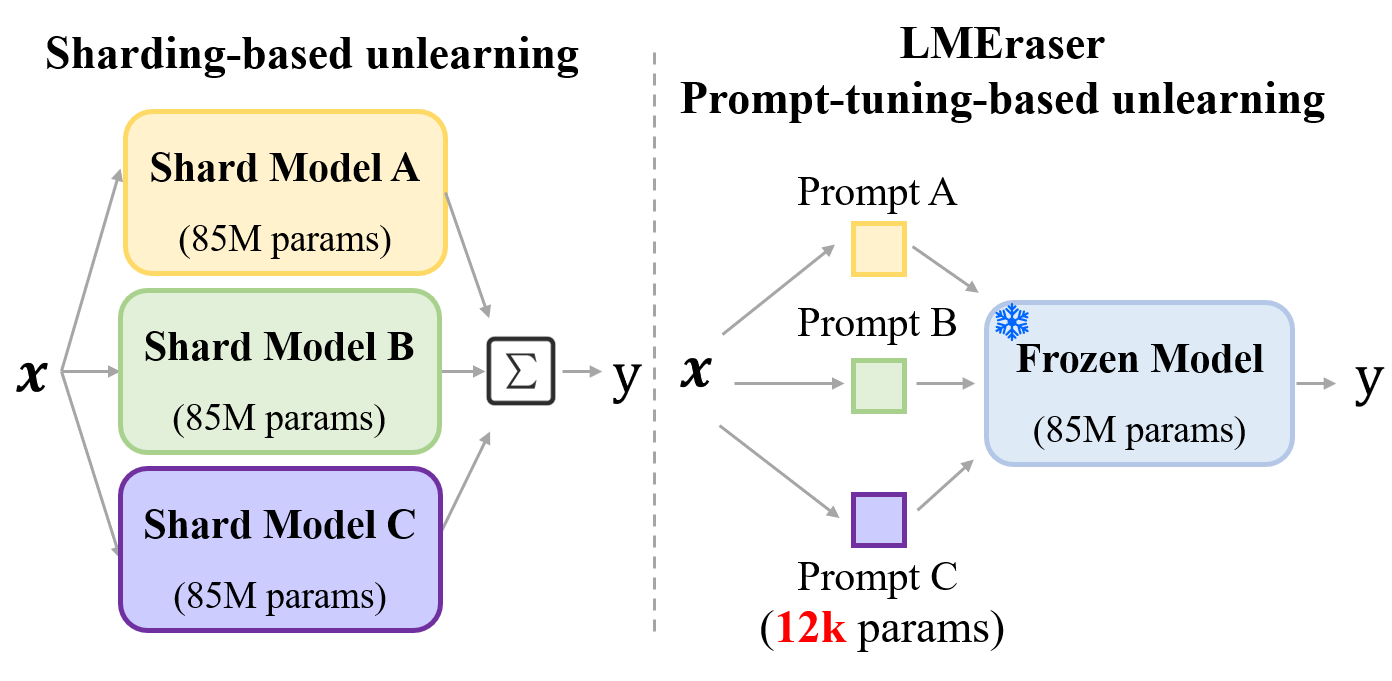}
	\caption{Comparing sharding-based unlearning and LMEraser. Sharding-based unlearning methods require retraining the affected shard model (around $85$M parameters), while LMEraser efficiently retrains the affected prompt (around $12$k parameters), significantly reducing retraining costs.}
	\label{fig:excompare}
\end{figure}

Existing machine unlearning methods are designed for smaller models and are inefficient and ineffective for large models. For example, some sharding-based methods divide the training dataset randomly into several shards~\cite{bourtoule2021,Brophy2020DaRE}. Each shard is used to train an independent model, and their collective outputs form the final prediction, as shown in Figure \ref{fig:excompare}. They improve unlearning efficiency by retraining the affected constituent shard models from scratch. It demands high computational and storage costs for large model unlearning, as multiple large constituent shard models must be trained and maintained.
Some studies update model parameters based on the estimated influence of removed data points on the model~\cite{guo2020certified,suriyakumar2022algorithms,wu2022puma}. However, these methods only achieve approximate unlearning and may disrupt the balance of learned parameters, potentially decreasing overall model performance. While existing methods can be useful in certain scenarios, they are not ideally suited for large models.

There are three key challenges of large model unlearning.
First, it is highly complex to \textit{identify} the influence of specific data points in large models. Large models are trained with massive datasets using batch learning. The influence of any single data point becomes highly dispersed across the entire parameter space of the model, making it difficult to isolate the data influence. 
Second, even with the ability of identifying data influences, \textit{implementing} the unlearning process is computationally expensive. Large models have complex architectures with billions of parameters. Unlearning operations involve intensive recalculations and adjustments within the model, demanding massive computational resources.
Third, it is difficult to maintain overall model performance when removing specific learned data points. Large models distribute learned representations across layers, with lower layers influencing the upper ones. Unlearning one data point can disrupt the model's knowledge of other data points, leading to performance degradation or even catastrophic forgetting~\cite{nguyen2020variational}. 

In this paper, we propose LMEraser, an efficient, exact unlearning method that can remove the influence of data points on large models without full retraining. The fundamental idea of LMEraser lies in two observations: 
1) The training data of large models can be divided into two categories: \textit{public data} and \textit{private data}. The public data come from open internet resources or non-profit organizations such as ImageNet~\cite{deng2009imagenet} or Wikipedia~\cite{Wikipedia2024}. The private data are sensitive and require permissions for access and use, such as medical records or bank transactions. While users (private data owners) may request that their data be removed from trained models due to the ``right to be forgotten'', such concerns are non-existent with public data. 2) Large models adopt a two-phase training process: the first pre-training phase trains a backbone using massive data that absorbs wide-ranging foundational knowledge. The second tuning phase utilizes the backbone and downstream data to enable the model to handle target tasks. 

Based on these observations, LMEraser has three core designs to address the challenges of large model unlearning. 
First, LMEraser takes a divide-and-conquer strategy with a prompt tuning architecture to isolate data influence. The training dataset is partitioned into public and private datasets. Public data are used for pre-training the model's backbone, and private data are used for prompt tuning.
Second, LMEraser freezes the backbone's parameters after pre-training. It reduces the computational costs for recalculations during unlearning and prevents catastrophic forgetting to ensure the model's stability.
Finally, LMEraser takes an adaptive prompt tuning mechanism to balance unlearning efficiency and model performance. It adaptively clusters the private data according to their diversity and generates tailored prompts for each homogeneous cluster, enabling more precise learning and targeted unlearning. Predictions are made with the nearest cluster's prompt to ensure model robustness and accuracy. To remove a training data point, only the parts related to its cluster are retrained, while the backbone and other parts remain unchanged.

Overall, the contributions are summarized as follows.
\begin{enumerate}[label=\arabic*)]
	\item We propose LMEraser, a pioneering approach to machine unlearning for large models. LMEraser takes a divide-and-conquer strategy to isolate private data influences from public data using a prompt tuning architecture. 
	\item We design an adaptive prompt tuning mechanism that includes adaptive private data clustering and tailored prompt generation, minimizing unlearning costs and maintaining model performance.
	\item We conduct extensive experiments to evaluate LMEraser's performance on image classification tasks. The results show that, compared to baseline methods, LMEraser achieves a $100$-fold reduction in unlearning costs while still maintaining high accuracy.
\end{enumerate}

\section{Background and Related Work}
\subsubsection{Prompt Tuning}
Prompt tuning provides an efficient way to adapt large pre-trained models to new downstream tasks by appending small learnable vectors, or ``prompts," to the input data, eliminating the need to update model parameters~\cite{liu2023pre}. It reformulates downstream data into the format of the pre-training data, allowing the model to utilize the knowledge from its original extensive training~\cite{huang2023diversity}. This method is computationally and storage efficient, as a single model can handle multiple tasks with different prompts~\cite{lester2021power}. 

Visual Prompt tuning often utilizes pre-trained Vision Transformers (ViT)~\cite{dosovitskiy2020image}, which process images by dividing them into patches and converting these into token embeddings. VPT~\cite{jia2022visual} integrates learnable prompts as embeddings alongside image embeddings in a pre-trained model, while VP~\cite{bahng2022exploring} utilizes pixel-level prompts similar to photo frames. DAM-VP~\cite{huang2023diversity} addresses the oversight of dataset diversity in VP and VPT, proposing a diversity-aware method, but its head-missing approach shows poor performance. APT~\cite{bowman2023carte} constructs user-centric models with prompts generated from user-selected sub-datasets, but it overlooks the importance of data partitioning, adversely affecting model performance.

\begin{figure*}[ht]
	\centering
	\includegraphics[width=0.9\linewidth]{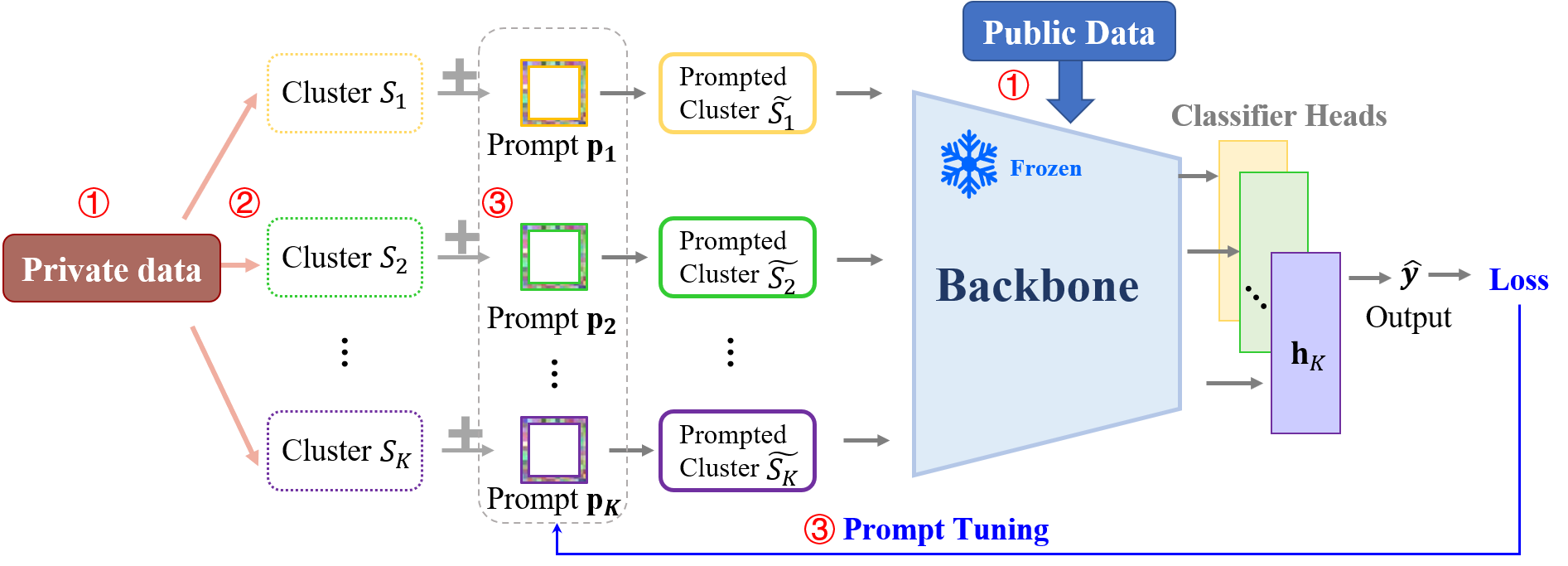}
	\caption{Overview model training process of LMEraser. (1) Partitioning training data into public and private datasets and pre-training backbone on public data. (2) Clustering private data based on diversity. (3) Tuning prompts and classifier heads for each cluster. }
	\label{fig:overview}
\end{figure*}

\subsubsection{Machine Unlearning} 
Machine unlearning enables the removal of specific data from trained models without full retraining, critical for addressing user privacy concerns~\cite{cao2015towards}.

 \textit{Exact unlearning} methods completely remove a data point's influence on the model~\cite{liu2022right}. This typically involves dividing the training data into shards, training separate models on these shards, and retraining only the affected models when data is unlearned. While effective, this process is resource-intensive, requiring substantial time and computational resources~\cite{bourtoule2021,Brophy2020DaRE,su2023asynchronous}. 
 
\textit{Approximate unlearning} aims to efficiently reduce the influence of data points to an acceptable level~\cite{xu2023unlearning}. Some studies use influence functions~\cite{Koh2017Understanding} to estimate and adjust for the influence of the removed data~\cite{guo2020certified,sekhari2021remember,suriyakumar2022algorithms,wu2022puma,warnecke2023machine,wu2023gif}. However, these methods rely on convex loss functions and inverting the Hessian matrix, which is infeasible for large models. Other methods, such as noise injection and selective retraining~\cite{golatkar2020eternal,golatkar2020forgetting,golatkar2021mixed}, model pruning~\cite{wang2022federated} have also been explored. Despite these advancements, Thudi \textit{et al.} ~\cite{thudi2022necessity} argues that approximate unlearning methods cannot provide unlearning proof, so they cannot address users' privacy concerns. 

\section{Preliminaries and Problem Definition}
\subsection{Vision Transformer (ViT)}
The Vision Transformer (ViT)~\cite{dosovitskiy2020image} splits an input image $\mathbf{x} \in \mathbb{R}^{H \times W \times C}$, where $H$, $W$, $C$ are height, width and channels, into non-overlapping patches. These patches are flattened into embeddings and combined with a learnable classification token to form a sequence $\mathbf{Z}$. $\mathbf{Z}$ is input to a Transformer encoder~\cite{vaswani2017attention}, whose output includes classification results.

\subsection{Visual Prompt Tuning}
\begin{definition}[Pixel-frame Visual Prompt]
	A pixel-frame visual prompt $\mathbf{p} \in \mathbb{R}^{H' \times W' \times C'}$ is a small perturbation overlayed on an image $\mathbf{x}$ as shown in Eq.(\ref{eqphi}).
	\begin{equation}\label{eqphi}
		\tilde{\mathbf{x}} = g(\mathbf{x}, \mathbf{p}),
	\end{equation}
	where $g: \mathbb{R}^{H \times W \times C} \times \mathbb{R}^{H' \times W' \times C'} \rightarrow \mathbb{R}^{H \times W \times C}$ is a function that overlays $\mathbf{p}$ onto $\mathbf{x}$. The prompted image $\tilde{\mathbf{x}}$ is then input into a pre-trained model.
\end{definition}

Visual prompt tuning optimizes $\mathbf{p}$ to minimize a task-specific loss function, resulting in an optimized prompt $\mathbf{p}^*$ that enhances the pre-trained model's performance on downstream tasks.

\subsection{Problem Definition}
Our goal is to design an exact machine unlearning method that can completely remove the influence of any private data point from a trained model when requested. Consider a trained machine learning model $f: X \rightarrow Y$ with parameters $\mathbf{\theta}^*$, trained on dataset $\mathcal{D} = \{(\mathbf{x}_i, y_i)\}_{i=1}^N$, where $\mathbf{x}_i \in X$ are inputs and $y_i \in Y$ are labels. The unlearning method aims to obtain an unlearned model $f': X \rightarrow Y$ with parameters $\mathbf{\theta}'$ that behaves as if it were trained on the dataset $\mathcal{D}' = \mathcal{D} \setminus \mathcal{D}_u $, where $\mathcal{D}_u \subseteq \mathcal{D}$ is the data to be removed. The unlearned $f'$ should have no residual influence from removed data, maintain overall performance, and enable efficient unlearning without full retraining.

\section{Method}
\subsection{Data Partitioning and Pre-training}
Large models introduce complexities in data management and pose challenges to privacy protection. Existing methods neglect differences in data sources and directly train large models on mixed \textit{public} and \textit{private} datasets~\cite{bourtoule2021}. This simplifies training but causes several problems. 
1) Privacy risks: Without data isolation, sensitive private data, when processed alongside public data, may be unintentionally exposed or misused, resulting in legal and ethical risks.
2) Inefficient unlearning: When users request data removal under regulations such as GDPR~\cite{Voigt2017}, the intertwined dependencies in mixed data sets complicate the process of machine unlearning, as each request necessitates retraining or re-optimizing the model using the entire training dataset.

To address these challenges, LMEraser partitions the training dataset $\mathcal{D}$ into a public dataset $\mathcal{D}_p$ and private dataset $\mathcal{D}_s$ based on their sources as shown in Eq. (\ref{eqsd}).
\begin{equation}\label{eqsd}
	\mathcal{D} = \mathcal{D}_p \cup \mathcal{D}_s, \text{and} \quad \mathcal{D}_p \cap \mathcal{D}_s = \emptyset.
\end{equation}
Public data are sourced from research institutions or non-profit organizations such as ImageNet~\cite{deng2009imagenet} and Wikipedia~\cite{Wikipedia2024}. 
In contrast, private data contain sensitive information, such as medical records and bank transactions, which require authorization for access and usage. 

LMEraser pre-trains a ViT backbone $\mathcal{M}$ solely using the public dataset $\mathcal{D}_p$ as shown in Figure \ref{fig:overview}. This approach ensures that the backbone acquires generalized visual representations by learning from diverse public images, avoiding any influence from sensitive private data. The pre-trained backbone's parameters $\mathbf{\theta}$ are frozen after pre-training. This step has two benefits. First, it prevents the influence of private data on the pre-trained model, thus eliminating the need for complete retraining during unlearning requests. Second, it enhances model stability by preserving the knowledge acquired during pre-training, mitigating the risk of catastrophic forgetting where removing a single data point could disrupt the learned knowledge of others~\cite{Shibata2021}. 

\subsection{Adaptive Prompt Tuning}
\subsubsection{Private Data Clustering Based on Diversity}
Although pre-training on public datasets equips models with foundational knowledge, it lacks domain-specific representations. In contrast, private data offer higher quality, unique features, and non-standard distributions, making it a rich source for domain-specific customization \cite{bommasani2021opportunities}. While beneficial for model generalization, the wide variability in data characteristics within private datasets introduces challenges in optimizing prompt tuning for enhanced model performance \cite{zhou2022domain}. Existing studies show that data diversity significantly affects the effectiveness of prompts~\cite{bahng2022exploring}. A single prompt may be sufficient for low-diversity datasets, but it becomes less effective for datasets with high diversity in features and patterns. Moreover, using just one prompt makes the process of unlearning less efficient. To remove a private data point, it is necessary to retrain this single prompt using all the private data.

To address these challenges, we propose an adaptive prompt tuning mechanism, which is inspired by DAM-VP's diversity-aware method~\cite{huang2023diversity} but uses multiple classifier heads for efficient large model unlearning. LMEraser first clusters private data adaptively based on their diversity and then trains tailored prompts and classifier heads for each cluster. 

Our adaptive prompt tuning mechanism offers two significant benefits:
1) Efficient unlearning. The influence of private data is constrained to their respective clusters. When unlearning is requested, only the prompts and classifier heads of the affected clusters require re-optimization rather than the entire model. This targeted approach minimizes unlearning costs and maintains model stability.
2) Enhanced feature recognition:  Clustering similar data allows cluster-specific prompts to detect nuanced differences, significantly boosting the model's feature recognition and pattern analysis capabilities. This approach contrasts with random sharding methods~\cite{bourtoule2021}, which randomly partition data into a predetermined number of shards without considering data characteristics. LMEraser's data-driven adaptive clustering offers enhanced adaptability in managing diverse data distributions, achieving higher accuracy, as demonstrated in Section \ref{secevaabla}.

Specifically, the adaptive clustering process initiates with a random sampling of the private dataset, extracting feature embeddings $\mathbf{f}_i$ for each data point $\mathbf{x}_i$ using the pre-trained model backbone $\mathcal{M}$. This sampling step expedites clustering for large datasets and improves unlearning efficiency, as re-clustering is necessary only when sampled data are targeted for unlearning. 
A hierarchical clustering algorithm~\cite{mullner2011modern} processes these embeddings, forming homogeneous clusters $\mathbf{S}_1,...,\mathbf{S}_K$ based on intrinsic data diversity. Prototype features $\mathbf{o}_k$ for each cluster encapsulate the characteristic features of the cluster. Subsequently, unsampled private data points  $\mathbf{x}_i \in \mathcal{D}_s \setminus \mathcal{D}_{\text{sample}}$ are assigned to the cluster with the nearest prototype feature vector. This adaptive clustering process captures the underlying diversity of the full private dataset and clusters them effectively. The detailed adaptive clustering process is shown in Algorithm \ref{alg_cluster}.

\begin{algorithm}[tb]
	\caption{Adaptive Private Data Clustering} 
	\label{alg_cluster}
	\textbf{Input:} Private dataset $\mathcal{D}_s$, pre-trained model backbone $\mathcal{M}$\\
	\textbf{Output:} Clustered private dataset $\{\mathbf{S}_1,...,\mathbf{S}_K\}$ with cluster prototypes $\{\mathbf{o}_1, \mathbf{o}_2,\ldots, \mathbf{o}_K\} $
	\begin{algorithmic}[1]
		\State $\mathcal{D}_{\text{sample}} \gets \text{RandomSample}(\mathcal{D}_s)$ 
		\For{each data point $\mathbf{x}_i$ in $\mathcal{D}_{\text{sample}}$}
		\State $\mathbf{f}_i \gets \mathcal{M}(\mathbf{x}_i)$ \Comment{Extract feature of $\mathbf{x}_i$}
		\EndFor
		\For{each pairs $(i, j)$ in $\mathcal{D}_{\text{sample}}$ }
		\State $\mathbf{M}_{ij} \gets \| \mathbf{f}_i - \mathbf{f}_j \|_2$ \Comment{Euclidean distance matrix}
		\EndFor	
		\State $ \{\mathbf{S}_1, \mathbf{S}_2,\ldots, \mathbf{S}_K\} \gets$ \Call{HierCluster}{$\mathbf{M},\mathcal{D}_{\text{sample}}$}		
		\For{each cluster $\mathbf{S}_{k}$}
		\State $\mathbf{o}_{k} \gets$ \Call{CompPrototype}{$\mathbf{S}_{k}$} \Comment{Prototype of $\mathbf{S}_k$}
		\EndFor
		\For{each data point $\mathbf{x}_i \in\mathcal{D}_s \setminus \mathcal{D}_{\text{sample}}$}
		\State $\mathbf{f}_{i} \gets \mathcal{M}(\mathbf{x}_i)$ \Comment{Extract feature of $\mathbf{x}_i$}
		\State $k \gets \arg\min_k \|\mathbf{f}_i - \mathbf{o}_k\|_2$ \Comment{Nearest prototype}
		\State Add $\mathbf{x}_i$ to cluster $\mathbf{S}_k$
		\EndFor
		\State \Return ${\mathbf{S}_1,...,\mathbf{S}_K}$, ${\mathbf{o}_1,...,\mathbf{o}_K}$
	\end{algorithmic}
\end{algorithm}

\subsubsection{Prompt Tuning for Clustered Private Data}
LMEraser uses pixel-frame prompts to enhance the pre-trained backbone model's recognition of private data. These prompts, small overlays that perturb the pixel values of an input image~\cite{bahng2022exploring}, are individually optimized for each private data cluster, thereby offering targeted enhancement that aligns with the unique distribution characteristics of that cluster.
Unlike traditional fine-tuning of model parameters, LMEraser optimizes prompts in parallel, significantly boosting computational efficiency ~\cite{lester2021power}. Furthermore, these pixel-frame prompts preserve the original dimensions of the input image and are agnostic to backbone architectures, ensuring broad applicability across various model types.

For each private data cluster $\mathbf{S}_k$, a unique prompt parameter $\mathbf{p}_k$ is initially assigned a random value. The random prompt overlay is then added to each data point $\mathbf{x}_i$ within cluster $\mathbf{S}_k$. By perturbing pixel values, the prompt adjusts the data distribution in cluster $\mathbf{S}_k$ to better match patterns learned by the backbone model during pre-training.

Subsequently, the training process involves optimizing a prompt $\mathbf{p}_k$ and classifier head parameter $\mathbf{h}_k$ for each cluster $k$. The objective is to minimize the sum of the cross-entropy loss function $L$, which measures the discrepancy between the predicted label of prompted input $\mathbf{x}_i + \mathbf{p}_k$ and the ground-truth label $y_i$ for each point $\mathbf{x}_i$ in cluster $\mathbf{S}_k$, as detailed in Eq. (\ref{eqoptph}).
\begin{equation} \label{eqoptph}
	\mathcal{L}_k(\mathbf{p}_k,\mathbf{h}_k) = \sum_{{(\mathbf{x}_i, y_i)}\in \mathbf{S}_k}L(H(\mathcal{M}(\mathbf{x}_i + \mathbf{p}_k;\mathbf{\theta} );\mathbf{h}_k), y_i)
\end{equation}
In this equation, $H(\cdot;\mathbf{h}_k)$ represents the classification function executed using the classifier head parameter $\mathbf{h}_k$.  

Through iterative gradient descent optimization,  the prompts adjust the data distribution in each cluster $\mathbf{S}_k$ to match the backbone's learned patterns better. This process enhances the model's recognition capabilities for prompted data points, leveraging its pre-existing knowledge. Simultaneously,  each classifier head $\mathbf{h}_k$  evolves into a specialized classifier for its respective prompted data cluster. This joint optimization of prompt and classifier head achieves tailored adaptation to the distinct characteristics of each data cluster. Crucially, the backbone parameters remain frozen during optimization to prevent catastrophic forgetting.

\subsection{Prompt-tuning-based Unlearning}
Traditional unlearning methods, such as full retraining and sharding-based approaches, are resource-intensive for large models. This is primarily due to the necessity of retraining the entire model to guarantee exact unlearning. The inefficiency severely limits the adaptability of large models in dynamic environments where unlearning requests are frequent.

LMEraser enables efficient and exact large model unlearning through adaptive prompt tuning. When a specific data point  $(\mathbf{x}_r, y_r)$ needs to be removed from cluster $k$, only the prompt $\mathbf{p}_k$ and classifier head $\mathbf{h}_k$ of that cluster are re-trained. The backbone and other components remain unchanged. The affected cluster $\mathbf{S}_k$ is updated to $\mathbf{S}'_k = \mathbf{S}_k \setminus {(\mathbf{x}_r, y_r)}$ by excluding the data point $(\mathbf{x}_r, y_r)$. The prompt $\mathbf{p}_k$ and head $\mathbf{h}_k$ are then re-optimized to minimize the loss to adapt to the revised data distribution of $\mathbf{S}'_k$, as illustrated in Eq. (\ref{eqregenpr}).
\begin{equation}\label{eqregenpr}
	\mathcal{L}'_k(\mathbf{p}_k ,\mathbf{h}_k ) = \sum_{(\mathbf{x}_i, y_i)\in \mathbf{S}'_k}L(H(\mathcal{M}(\mathbf{x}_i + \mathbf{p}_k;\mathbf{\theta} );\mathbf{h}_k), y_i)
\end{equation}

This targeted re-optimization ensures that the influence of the removed data point $(\mathbf{x}_r, y_r)$ is completely removed from the model. After re-optimization, the updated prompt and classifier head are integrated into the model, ensuring it accurately reflects the updated dataset while maintaining overall performance. By limiting re-optimization to the prompt and head of the affected cluster, LMEraser achieves both efficient and exact unlearning. It effectively removes the influence of removed data points without affecting the knowledge embedded in the frozen backbone and other unaffected components of the model.

\section{Evaluation} \label{seceva}
Our evaluation of LMEraser focuses on two key metrics: model utility and unlearning efficiency. Model utility, measured by the accuracy of image classification tasks, can verify that the core functionality of the model is not compromised during the unlearning process. Meanwhile, unlearning efficiency, measured by the time and computational costs required for unlearning requests, can assess the model's practicality in real-world applications.

\subsection{Experimental Setup}
\subsubsection{Public Dataset} Our experiment uses the ImageNet-22K dataset ~\cite{deng2009imagenet} as a public dataset due to its wide variety and volume. This dataset provides a comprehensive foundation for our model backbones, ensuring a broad representation of general image features.

\subsubsection{Private Datasets} 
We use smaller datasets as our private datasets to evaluate LMEraser performance in image classification tasks, including CIFAR10~\cite{krizhevsky2009learning}, CIFAR100~\cite{krizhevsky2009learning}, GTSRB~\cite{stallkamp2012man}, and SVHN~\cite{netzer2011reading}. These datasets present varying sizes, input dimensionalities, and class distributions, enabling a comprehensive evaluation across varying levels of task complexity. To standardize, images are resized to $256 \times 256$ and then cropped to $224 \times 224$ pixels. The specific characteristics of private datasets are summarized in Table ~\ref{tabdataset}.
\begin{table}[h]
	\centering
	\resizebox{0.99\columnwidth}{!}{%
		\begin{tabular}{cccc}
			\toprule
			Dataset & Dimensionality & \# Images& \# Classes \\
			\midrule
			CIFAR-10 & $32 \times 32 \times 3$ & 60,000 & 10 \\
			CIFAR-100 & $32 \times 32 \times 3$ & 60,000 & 100 \\
			SVHN & $32 \times 32 \times 3$ & 58,605 & 10 \\
			GTSRB & \begin{tabular}[c]{@{}c@{}}Variable (from $15 \times 15\times 3$ \\to $250 \times 250\times 3$) \end{tabular} & 21,312 & 43 \\
			\bottomrule
		\end{tabular}%
	}
	\caption{Statistical details of datasets.}
	\label{tabdataset}
\end{table}

\subsubsection{Pre-trained Models}
The Vision Transformer (ViT-B/16)~\cite{dosovitskiy2020image}, and Swin Transformer (Swin-B)~\cite{liu2021swin}, both pre-trained on ImageNet-22K are used as the backbone.

\subsubsection{Baselines} 
We compare the performance of LMEraser with several methods that can completely remove the influence of training data points from the model, highlighting its unique advantages as an exact unlearning method.
\begin{itemize}
	\item Retraining from Scratch: This method involves completely retraining the model, including the backbone and prompts, on the entire dataset for each unlearning request.
	\item SISA~\cite{bourtoule2021}: A state-of-the-art solution for exact unlearning in large models, this method partitions training data into shards and trains independent models for each shard. Only the affected shard is retrained for unlearning requests.
	\item Single prompt: This method is based on VP~\cite{bahng2022exploring}, using one prompt for the entire private dataset with a frozen public-data-trained backbone. Unlearning involves retraining only the prompt.
	\item DAM-VP~\cite{huang2023diversity}: This method clusters private data by diversity and generates a prompt for each cluster using an activation-based feature-to-label mapping technique. Unlearning involves retraining only the affected prompt. 
\end{itemize}

\begin{figure*}[htbp]
	\centering
	\begin{subfigure}[b]{0.24\textwidth}
		\includegraphics[width=\textwidth]{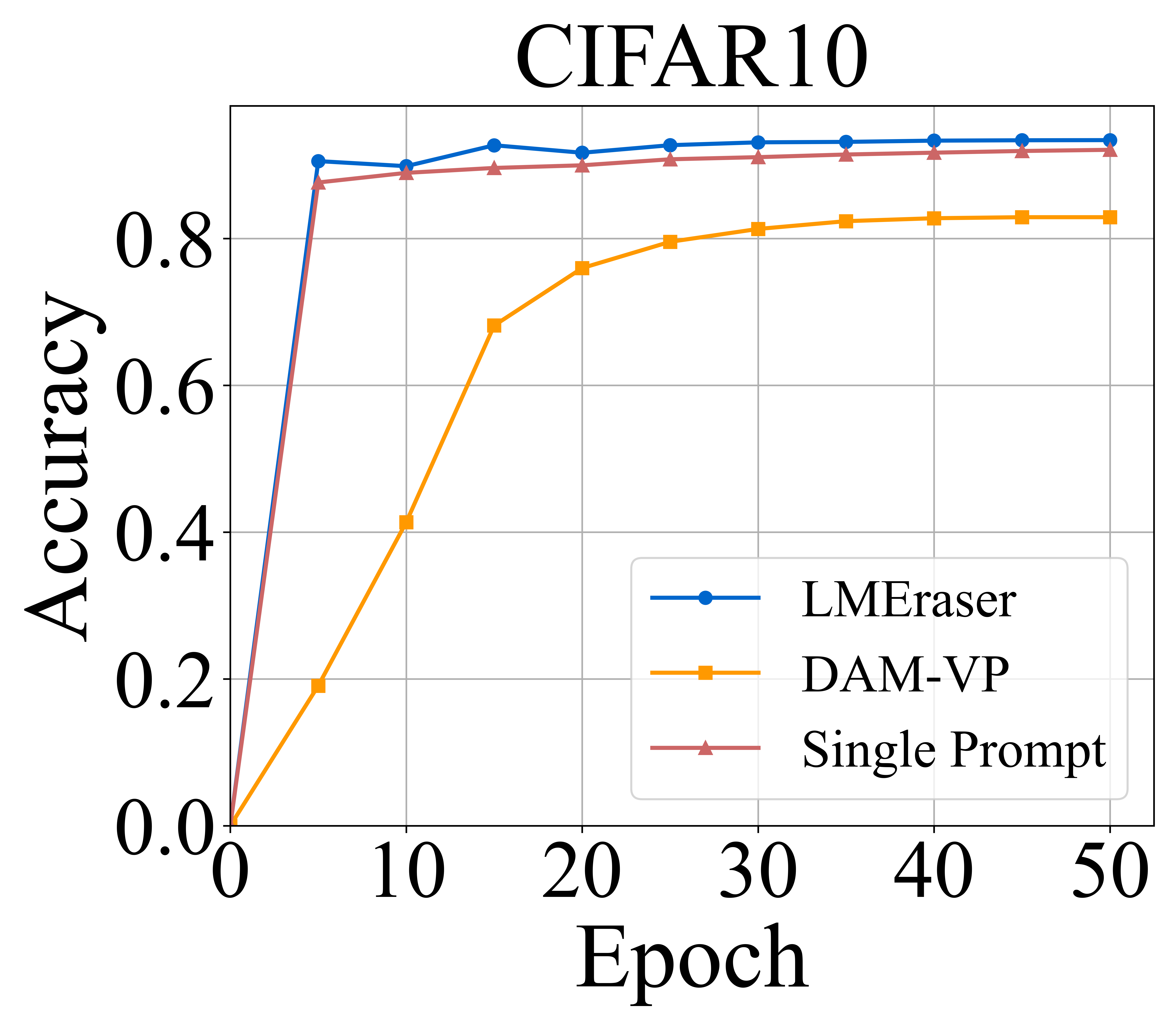}
		\label{fig:cmpbslsub1}
	\end{subfigure}
	\begin{subfigure}[b]{0.24\textwidth}
		\includegraphics[width=\textwidth]{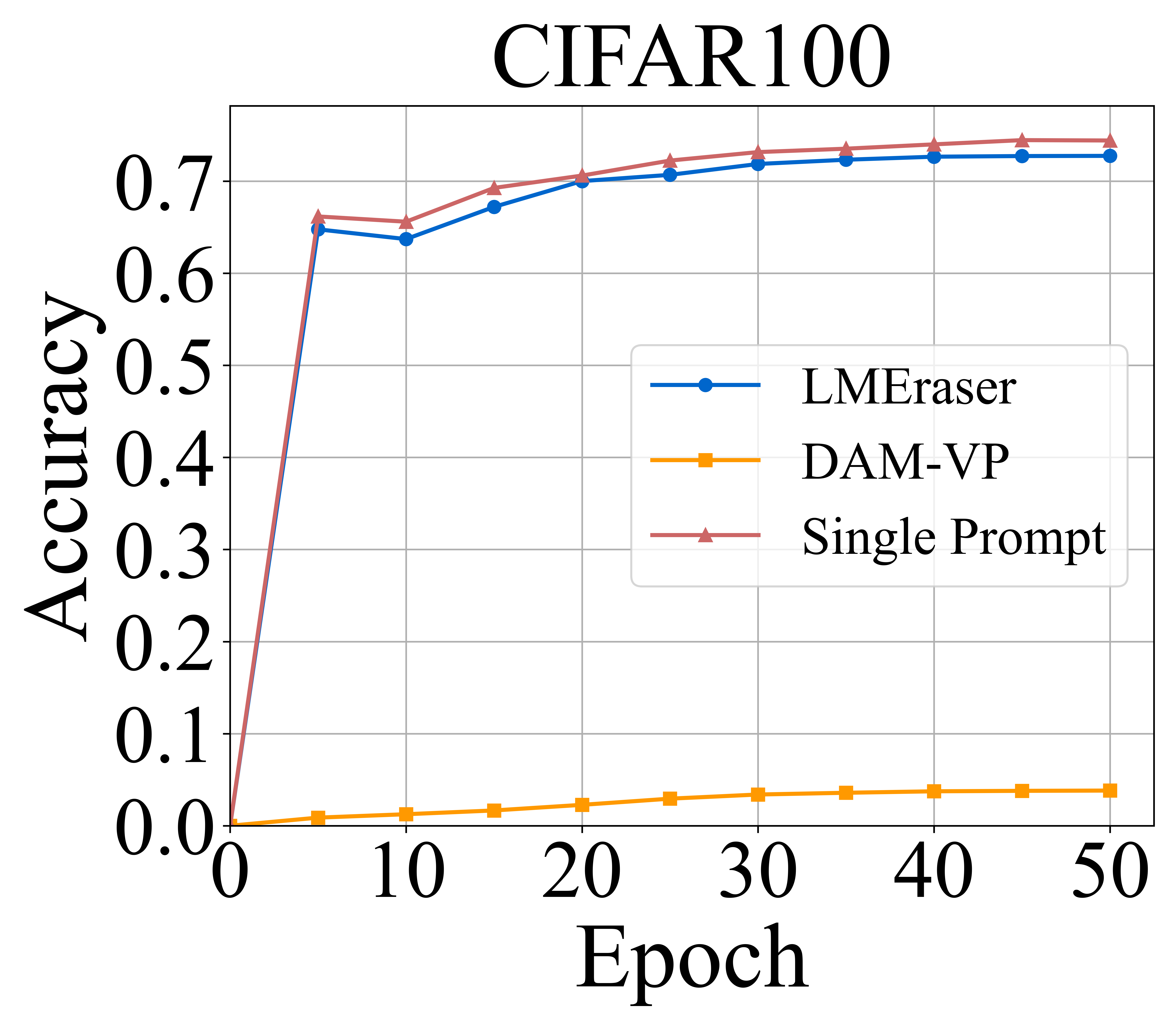}
		\label{fig:cmpbslsub2}
	\end{subfigure}
	\begin{subfigure}[b]{0.24\textwidth}
		\includegraphics[width=\textwidth]{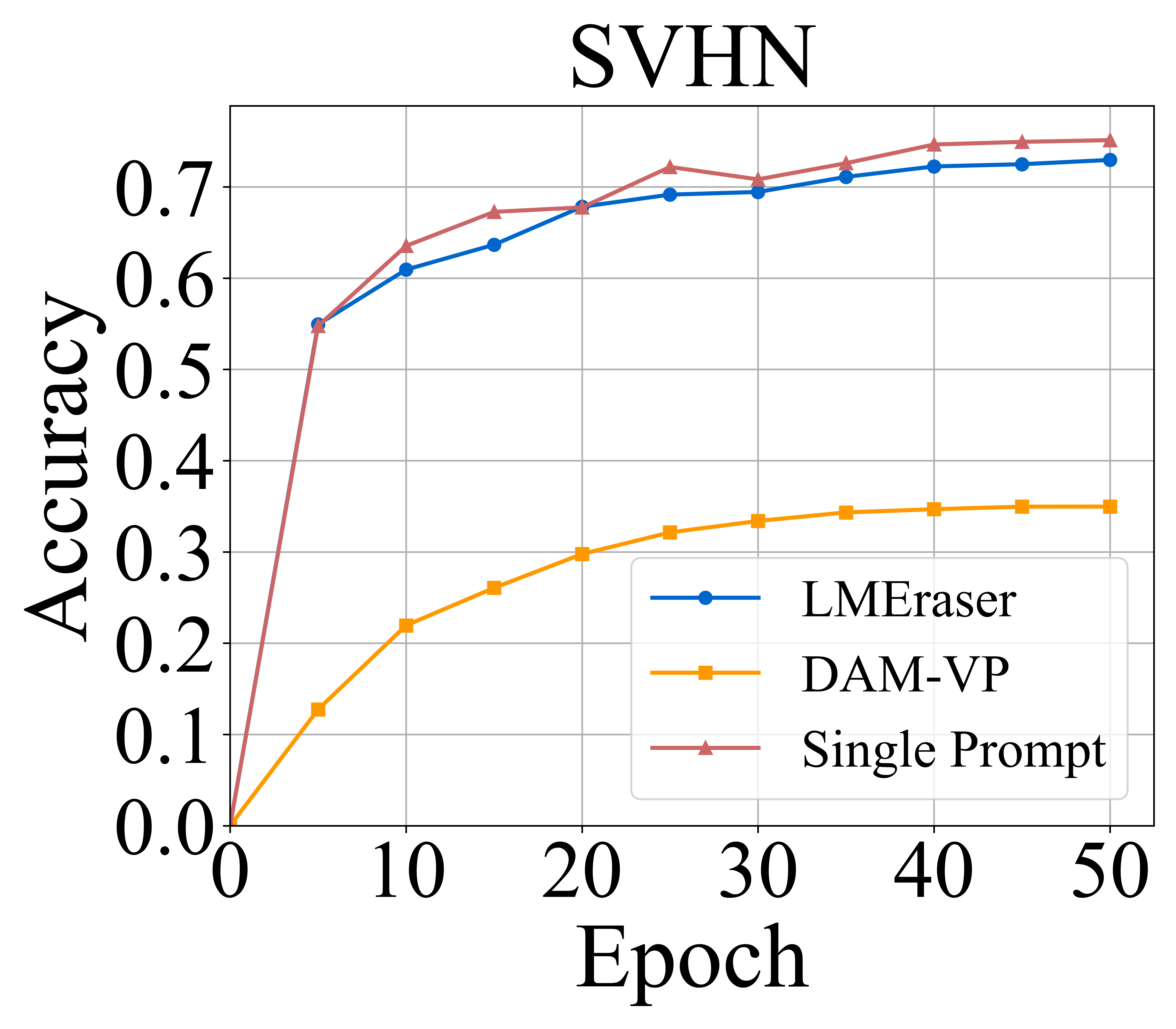}
		\label{fig:cmpbslsub3}
	\end{subfigure}
	\begin{subfigure}[b]{0.24\textwidth}
		\includegraphics[width=\textwidth]{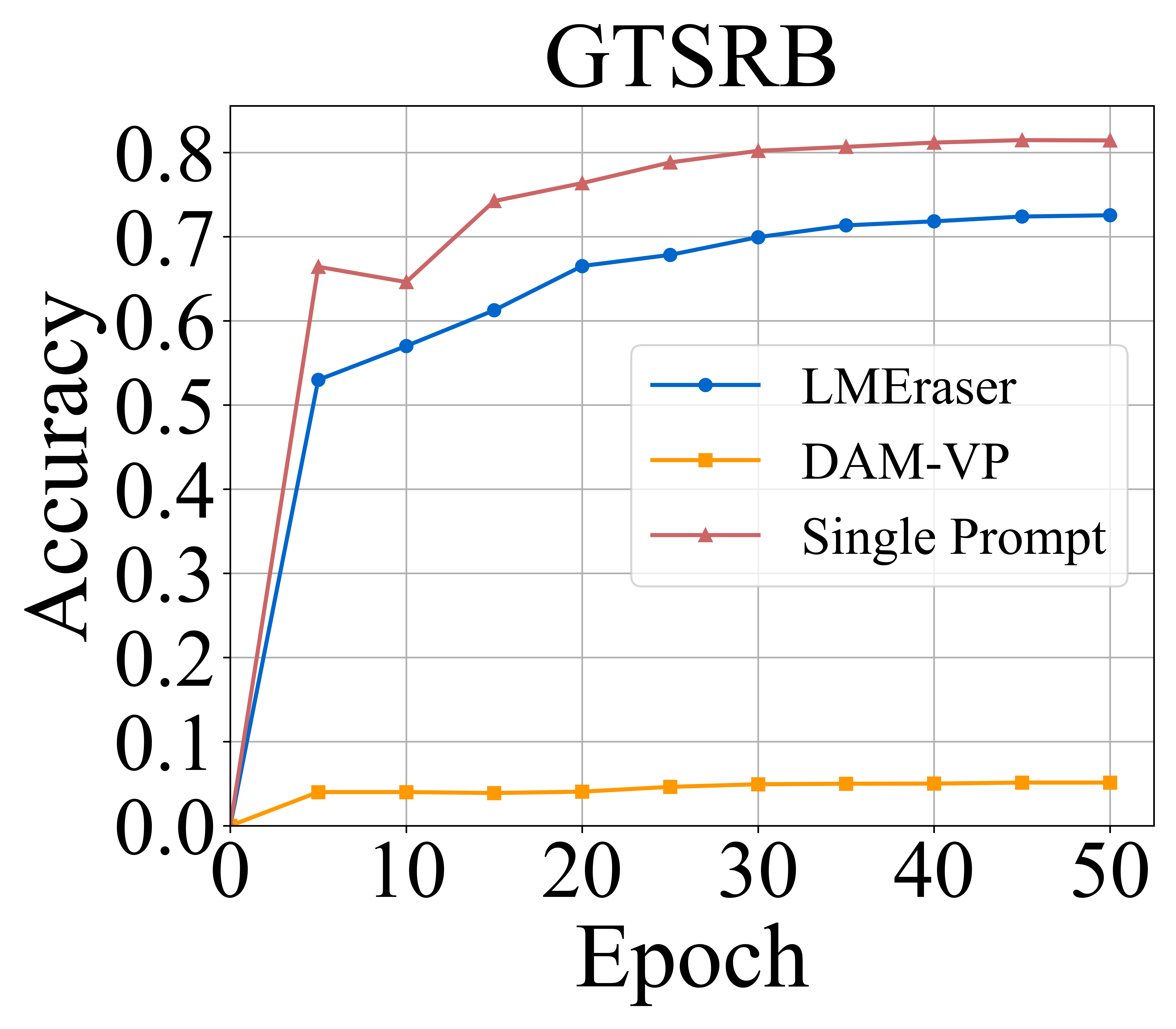}
		\label{fig:cmpbslsub4}
	\end{subfigure}
	\vspace{-5mm}
	\caption{Comparative analysis of test accuracy over epochs: LMEraser vs. baseline methods across various datasets  with ViT-B-22K.}
	\label{fig:cmpbslcombined}
\end{figure*}

\begin{figure*}[htbp]
	\centering
	\begin{subfigure}[b]{0.24\textwidth}
		\includegraphics[width=\textwidth]{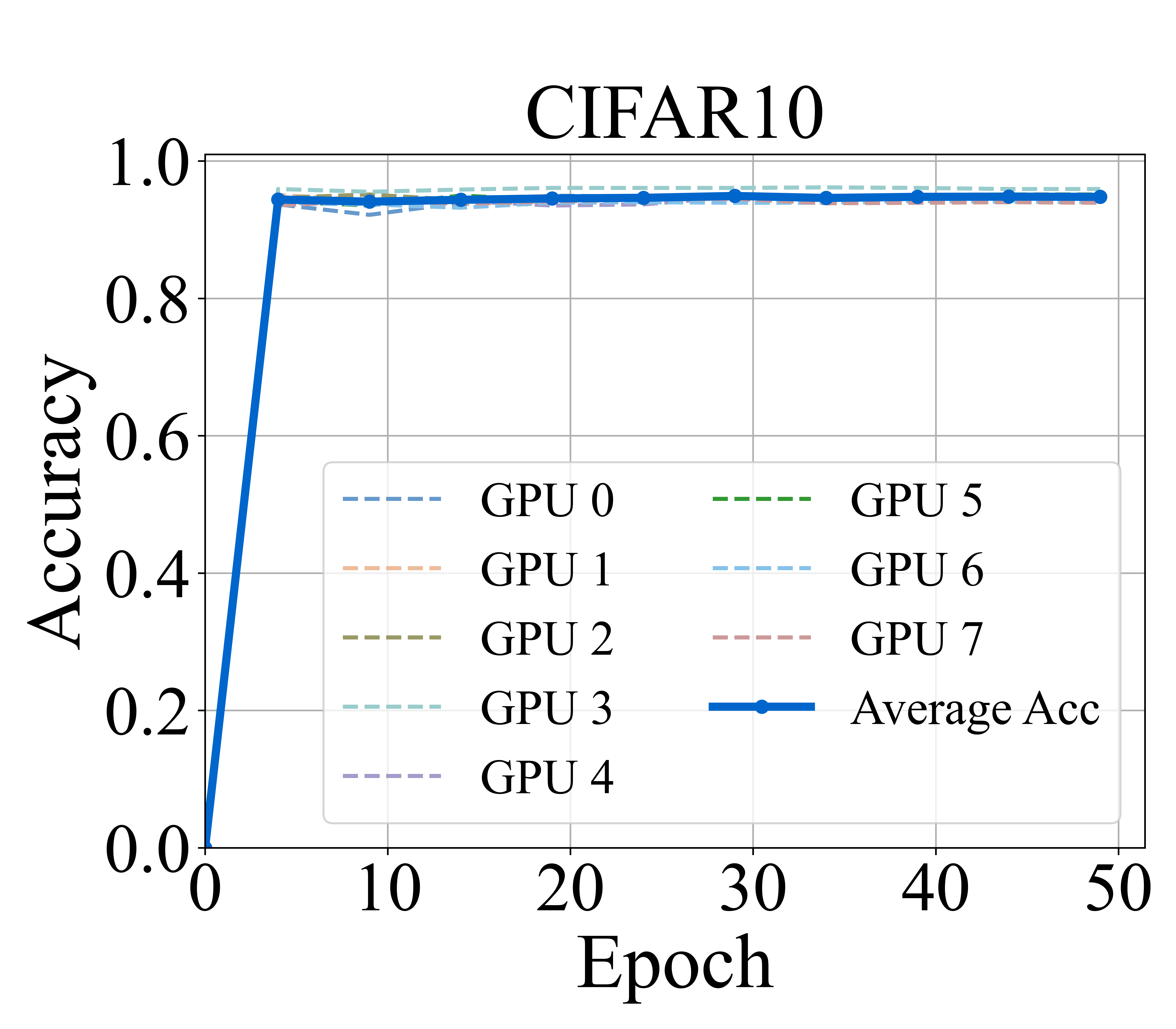}
		\label{fig:LMEraserSwinsub1}
	\end{subfigure}
	\begin{subfigure}[b]{0.24\textwidth}
		\includegraphics[width=\textwidth]{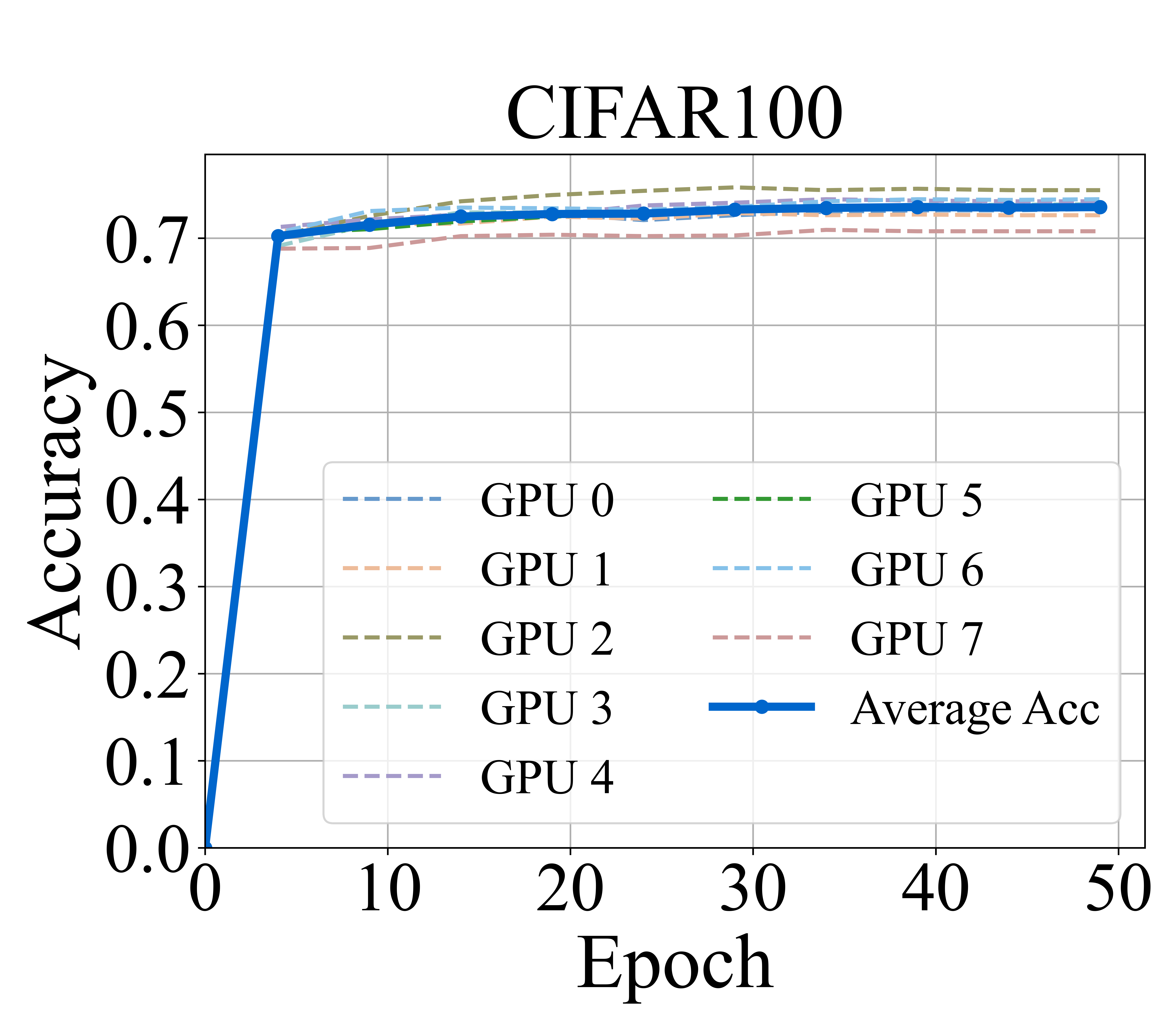}
		\label{fig:LMEraserSwinsub2}
	\end{subfigure}
	\begin{subfigure}[b]{0.24\textwidth}
		\includegraphics[width=\textwidth]{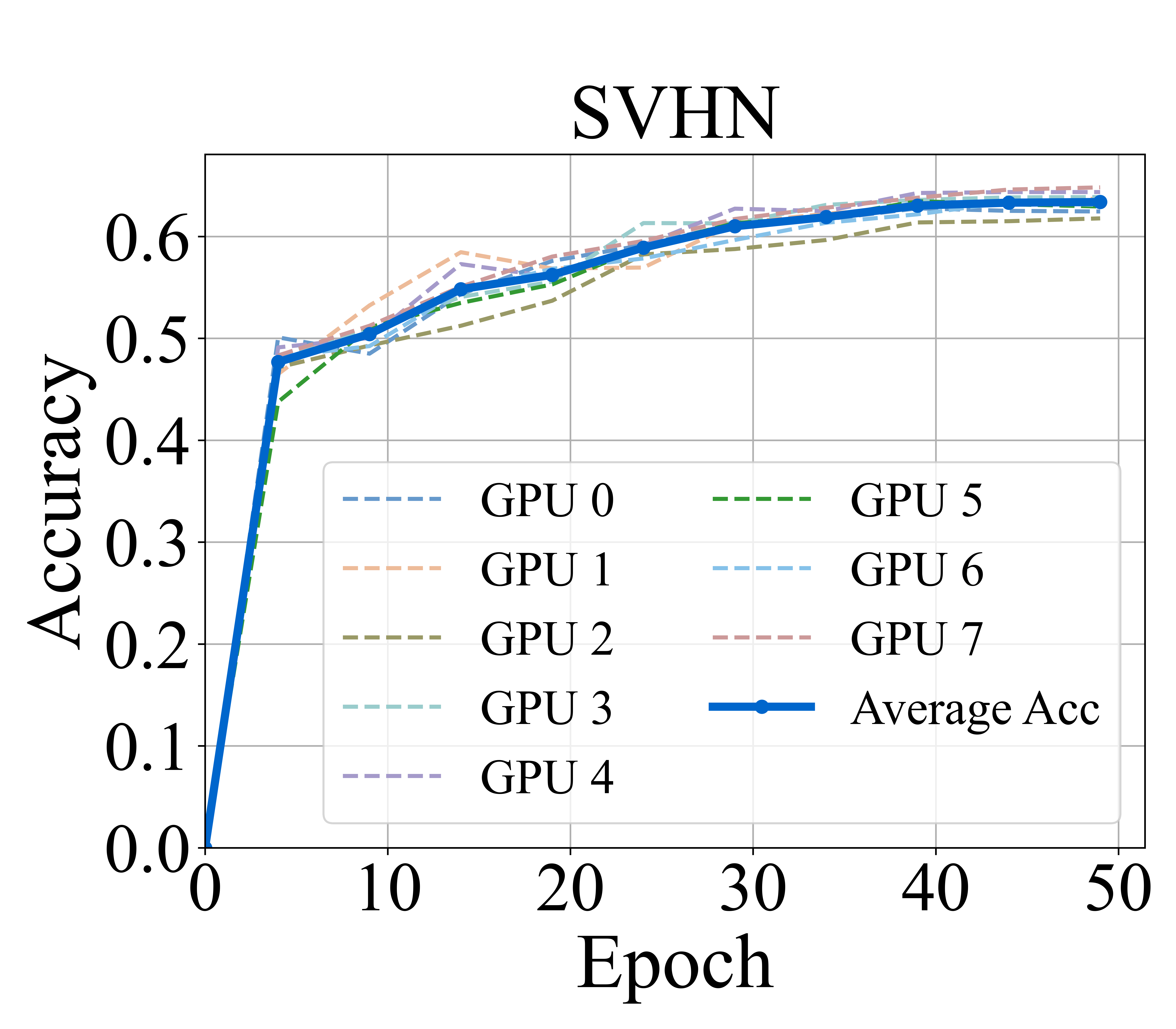}
		\label{fig:LMEraserSwinsub3}
	\end{subfigure}
	\begin{subfigure}[b]{0.24\textwidth}
		\includegraphics[width=\textwidth]{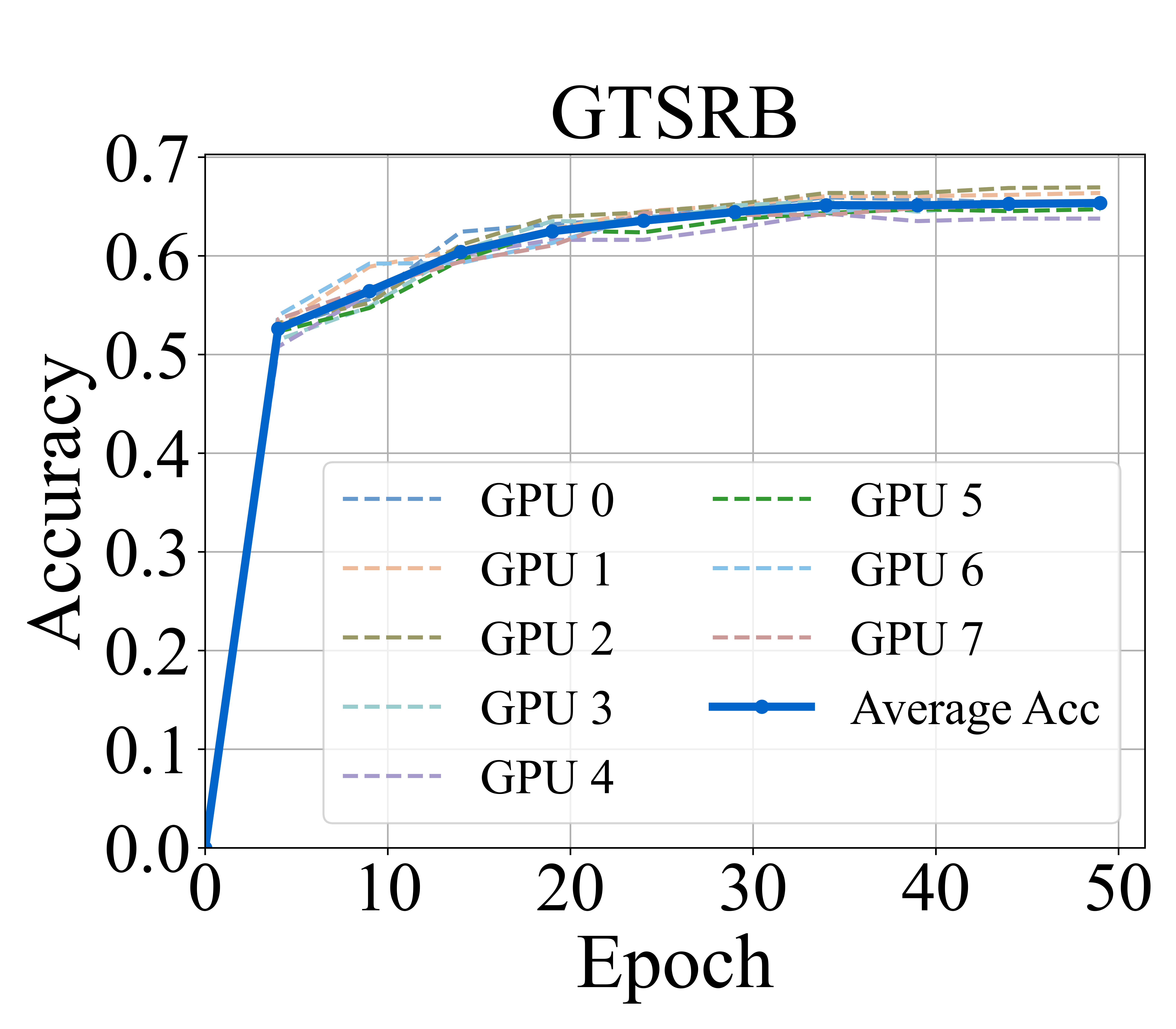}
		\label{fig:LMEraserSwinsub4}
	\end{subfigure}
	\vspace{-5mm}
	\caption{Performance evolution of LMEraser with Swin-B-22K: Test accuracy and average accuracy trends across epochs in 8 GPUs.}
	\label{fig:combined}
\end{figure*}

\subsubsection{Implementations}
Experiments are conducted using Nvidia Tesla V100-FHHL GPUs, each with 16 GB memory, and Intel Xeon CPU E5-2650 v4 @ 2.20GHz with 48 cores and 251GB of RAM. The setup runs on Ubuntu 22.04.3 LTS (64-bit), using PyTorch v2.1.2, CUDA 12.1, and Python 3.11.5.

Given platform constraints, we employ publicly available pre-trained parameters for ViT-B-22K and Swin-B-22K, reflecting real-world scenarios where training large models from scratch may be impractical. To evaluate private dataset diversity, the LPIPS metric \cite{zhang2018unreasonable} is applied, with clustering thresholds~\cite{mullner2011modern} adjusted for data diversity sensitivity. We use the same pixel-frame prompts as DAM-VP~\cite{huang2023diversity}, framing each image in the private dataset with a 30-pixel width.

\subsection{Evaluation of Model Utility}
 LMEraser takes an adaptive prompt tuning mechanism, where private datasets are clustered based on diversity. Each cluster is used to train a prompt and classifier head. Details of the clustering process and results can be found in Appendix \ref{apphiercluster}. We set the distance threshold for clustering at 10, which results in the number of clusters (and prompts) for CIFAR-10, CIFAR-100, SVHN, and GTSRB as 194, 347, 30, and 90, respectively. 
 
\subsubsection{Classification Task Accuracy Comparison}
We assess the test accuracy of LMEraser with ViT-B-22K  over the first 50 epochs, comparing it with baseline methods, as shown in Figure \ref{fig:cmpbslcombined}. 
LMEraser shows superior performance on all datasets, achieving comparable accuracy to the single prompt method and significantly outperforming DAM-VP, especially in tasks with a higher number of classes. This underscores LMEraser's effectiveness in managing diverse and complex data scenarios without accuracy compromise. 

\subsubsection{Performance with a Different Pre-trained Backbone} 
We evaluate the performance of LMEraser with Swin-B-22K backbone, monitoring test accuracy and average accuracy trends over epochs in eight identical GPUs. The results, depicted in Figure \ref{fig:combined}, indicate that LMEraser's performance with Swin-B-22K closely aligns with the outcomes observed using ViT-B-22K, as shown in Figure \ref{fig:cmpbslcombined}. The results demonstrate high test accuracy within the initial 20 epochs and stable convergence. This finding confirms LMEraser's efficiency and reliability in large-scale machine learning applications.

\begin{figure}[t]
	\centering
	\begin{minipage}{0.48\columnwidth}
		\includegraphics[width=\linewidth]{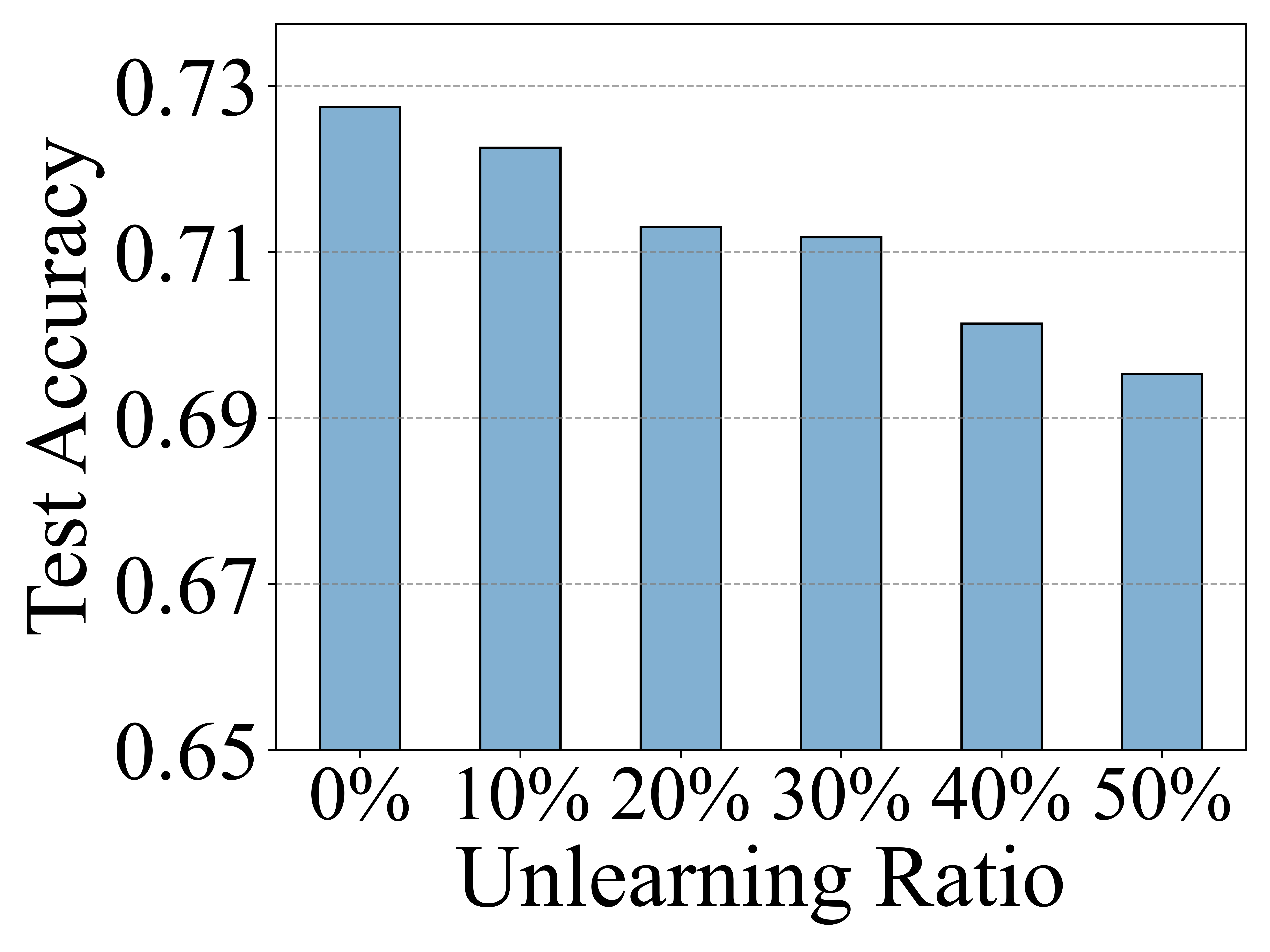}
	\end{minipage}\hfill
	\begin{minipage}{0.48\columnwidth}
		\includegraphics[width=\linewidth]{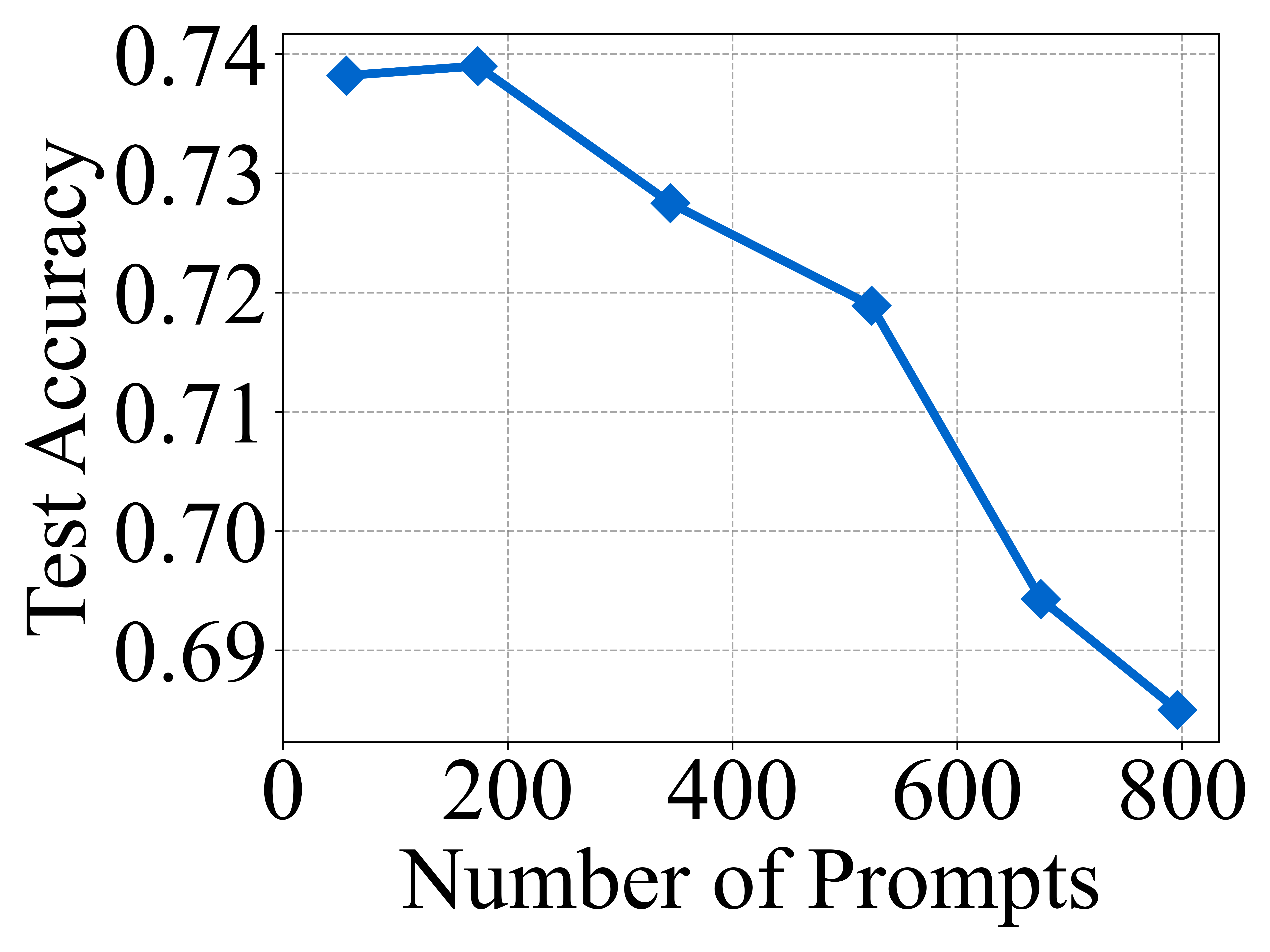}
	\end{minipage}
	\caption{LMEraser's test accuracy with various unlearning privacy data ratio and number of prompts (with ViT-B-22K backbone on CIFAR100 private dataset).}
	\label{fig:dynamicAcc}
\end{figure}

\subsubsection{Influence of Unlearning Ratio on Test Accuracy}  As illustrated in Figure \ref{fig:dynamicAcc}, a marginal decrease in test accuracy, from 0.7275 to 0.6953, occurs when the unlearning ratio increases from 0\% to 50\%.
This slight decline signifies LMEraser's robustness in data unlearning while maintaining high accuracy levels. 

\subsubsection{Effect of Prompt Quantity on Test Accuracy.} As shown in Figure \ref{fig:dynamicAcc}, the increase in the number of prompts, while reducing unlearning costs, adversely affects test accuracy. This observation highlights the importance of optimizing the balance between model utility and unlearning efficiency.

\subsection{Evaluation of Unlearning Efficiency}
\subsubsection{Comparing Affected Training Data and Model Size.} 
\begin{figure} [t]
	\centering
	\includegraphics[width=0.99\linewidth]{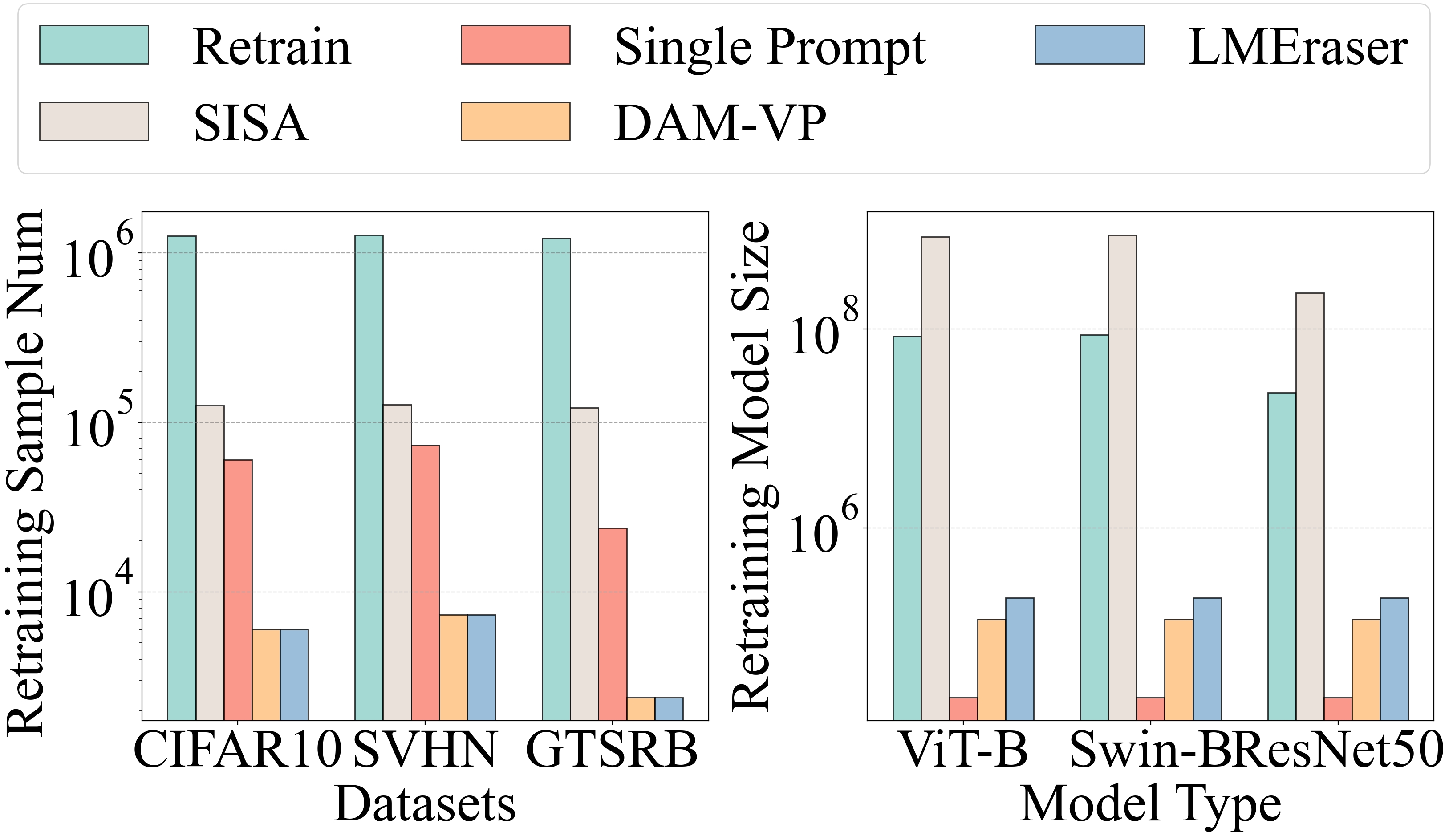}
	\caption{Comparison of affected training data points and model parameters that need to be retrained when a data point is unlearned in LMEraser and baseline methods (log scale).}
	\label{fig:compare2}
\end{figure}

The size of the affected training data and model parameters can directly reflect the unlearning costs. We assess the unlearning costs of LMEraser compared to baseline methods by examining the number of training data points and model parameters that require retraining to handle an unlearning request. In scenarios with ten shards or clusters, LMEraser demonstrates significantly lower unlearning costs than both naive retraining and the state-of-the-art solution, SISA, as shown in Figure \ref{fig:compare2}. Specifically, the affected training data points and model parameters in SISA and naive retraining are respectively $100$ and $1,000$ times greater than in LMEraser. 

\begin{table}[t]
	\centering
	\resizebox{0.99\columnwidth}{!}{%
		\begin{tabular}{c c c c c c c c c}
			\toprule
			Time & \textbf{Retrain}& \textbf{SISA}&\textbf{LMEraser}& \textbf{Single Prompt} & \textbf{DAM-VP}\\ \midrule
			CIFAR10 & Days & Days & 27.61s & 3185s& 21.26s  \\ 
			CIFAR100 & Days &Days & 12.45s& 4048s & 12.14s  \\ 
			SVHN & Days &Days & 213.81s& 4688s & 260.07s  \\ 
			GTSRB & Days &Days & 27.73s& 1740s& 26.64s  \\
			\bottomrule
		\end{tabular}
	}
	\caption{Comparison of the average time required for handling an unlearning request with backbone ViT-B-22K.}
	\label{tabcmpreprompt}
\end{table}

\subsubsection{Unlearning Time Costs}
We compare the average unlearning time of LMEraser with other methods using the ViT-B-22K backbone. The results are presented in Table \ref{tabcmpreprompt}. Retraining and SISA, requiring days to retrain large models, are impractical. LMEraser shows a great efficiency advantage, largely due to its ability to train multiple prompts in parallel. Unlike the single prompt method, LMEraser and DAM-VP can handle multiple unlearning requests simultaneously, demonstrating rapid adaptation and reduced computational costs across different datasets.

\subsection{Ablation Study}\label{secevaabla}
To assess LMEraser's core components, we conduct two ablation studies, focusing on the contributions of adaptive prompt tuning and multiple classifier heads to its performance and efficiency.

\begin{figure} [t]
	\centering
	\includegraphics[width=0.99\linewidth]{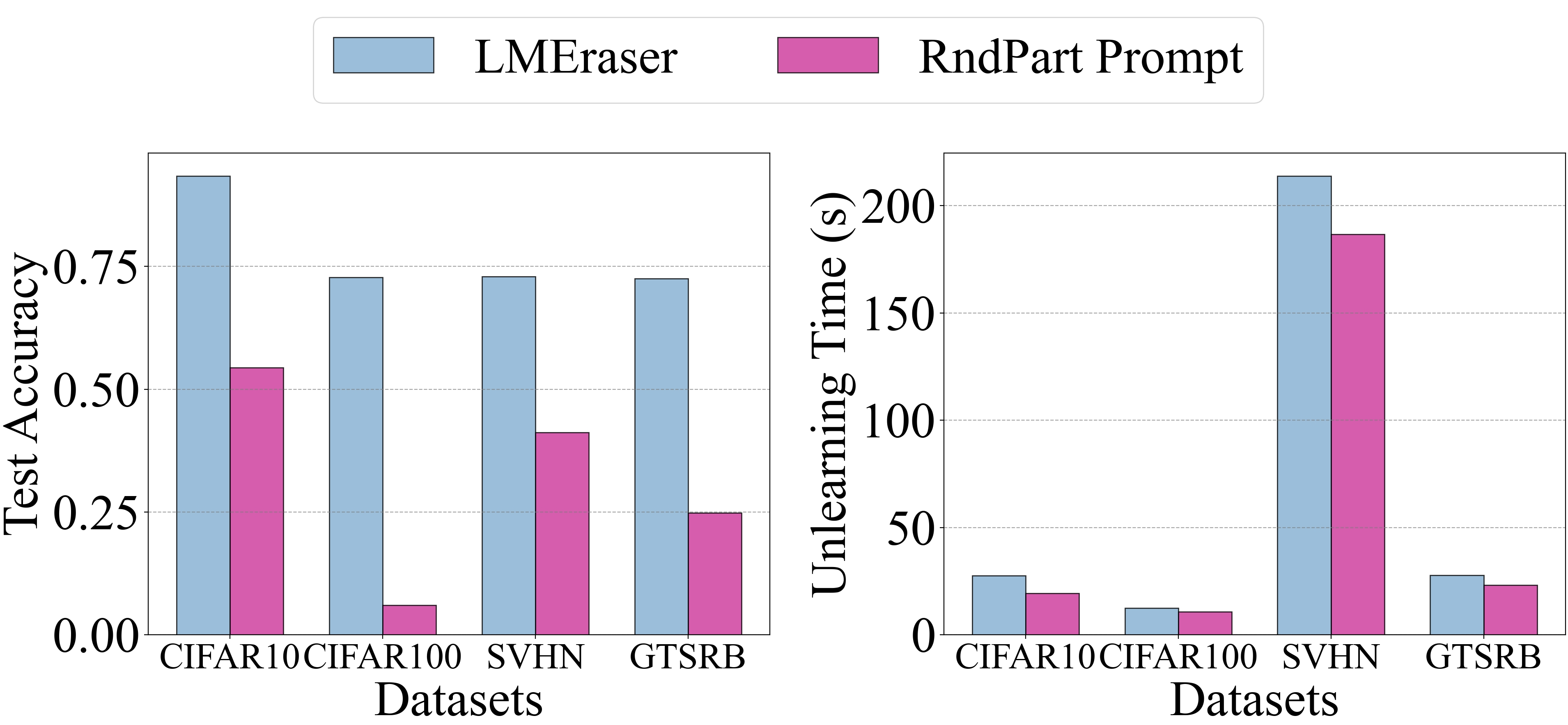}
	\caption{ Performance comparison: Random private data partitioning vs. LMEraser's adaptive clustering.}
	\label{figablationrand}
\end{figure}
\begin{figure} [t]
	\centering
	\includegraphics[width=0.99\linewidth]{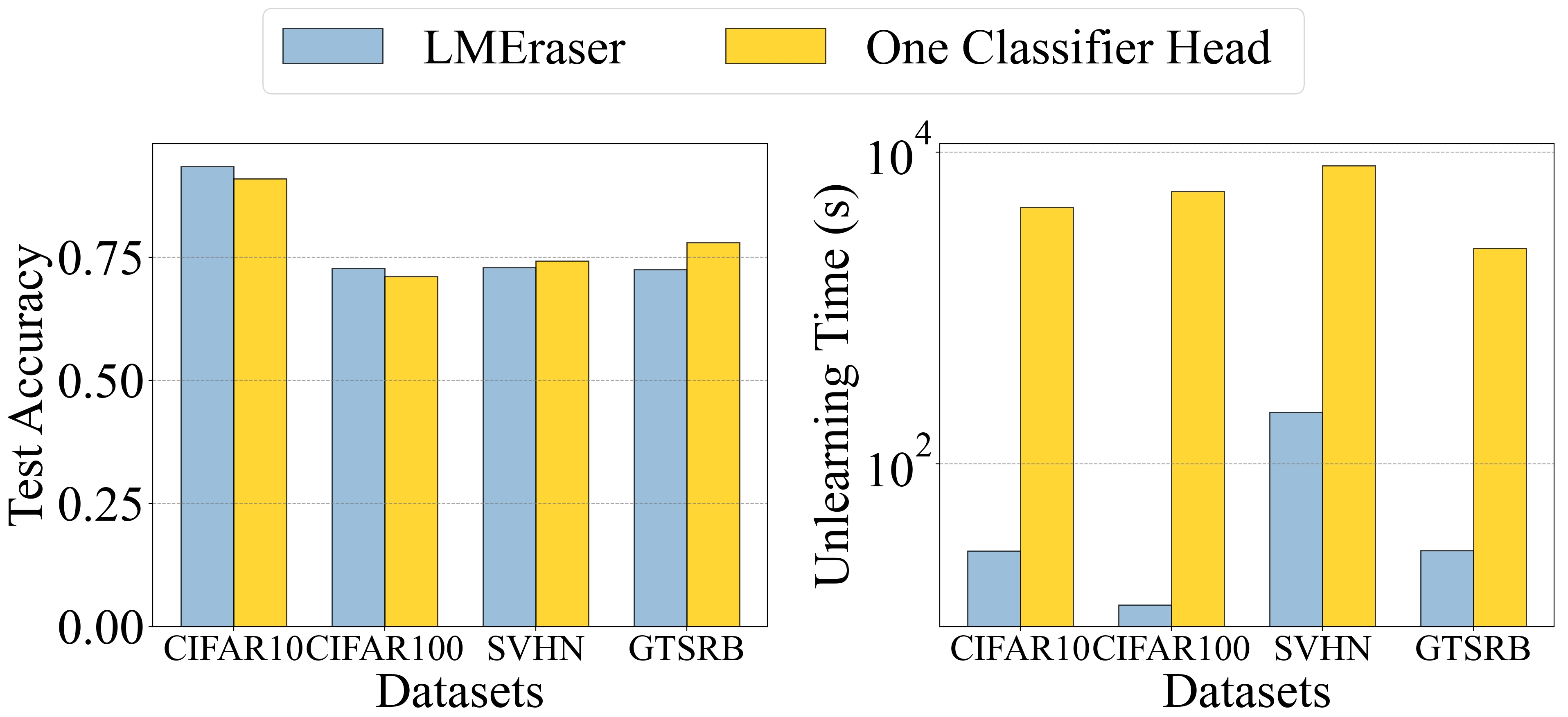}
	\caption{Performance comparison: One classifier head vs. LMEraser's multiple classifier heads.}
	\label{figablationMultiheads}
\end{figure}

\subsubsection{Adaptive Prompt Tuning}
We assess the effectiveness of adaptive clustering compared to a random partitioning approach (RndPart Prompt) in private data partitioning using a ViT-B-22K backbone, while maintaining identical other settings. Figure \ref{figablationrand} shows that both methods demonstrate comparable unlearning efficiency. However, the random partitioning approach performs poorly in tasks with a wide range of classes. This lower performance is primarily attributed to random partitioning's failure to account for the dataset's inherent diversity and distinct characteristics. LMEraser's approach of diversity-aware clustering and adaptive prompt tuning aligns the prompts more closely with the inherent characteristics of each data cluster, thereby significantly enhancing accuracy across various datasets.

\subsubsection{Multiple Classifier Heads} 
We compare the performance of using one classifier head against LMEraser's multiple heads approach, all while maintaining other conditions constant. Figure \ref{figablationMultiheads} shows that both methods exhibit similar levels of accuracy across different datasets. However, LMEraser's method stands out in unlearning efficiency. The one classifier head method, requiring retraining for each unlearning request, greatly increases the time and computational resources needed to fully remove the influence of a data point. In contrast, LMEraser's multiple classifier heads approach allows for targeted retraining of only the affected heads, significantly enhancing unlearning efficiency.  

\section{Conclusion}
This paper presents LMEraser, a pioneering, exact unlearning method designed for large models. LMEraser utilizes a prompt tuning architecture, effectively isolating the influence of private data points. Its adaptive prompt tuning method balances unlearning efficiency and preservation of model utility. Our comprehensive experiments demonstrate LMEraser's capability in achieving efficient unlearning without compromising accuracy, showcasing its adaptability across various datasets and large model architectures. 

\appendices
\section{Hierarchical Agglomerative Clustering} \label{apphiercluster}
\subsection{Hierarchical Agglomerative Clustering Algorithm} 
LMEraser employs hierarchical agglomerative clustering (HierCluster)~\cite{mullner2011modern} for partitioning the private dataset. By clustering private data, the influence of each data point is limited to its respective cluster, thereby reducing unlearning costs. Additionally, similar private data points are grouped within the same cluster, enabling cluster-specific prompts to detect subtle differences and enhance the model's utility.

HierCluster stands out for its efficiency and adaptability, which are crucial for effective unlearning. It handles diverse datasets flexibly and does not require pre-defining the number of clusters. This adaptability optimizes LMEraser's unlearning by allowing precise data clustering based on inherent characteristics, minimizing the impact on the model's performance while ensuring targeted data removal.

The HierCluster algorithm iteratively merges the most similar clusters based on their features, continuing until the distance between any two clusters exceeds the threshold or a single comprehensive cluster is formed. The threshold is determined based on the dataset's characteristics and the desired number of clusters. Optimal threshold values and their selection criteria are detailed in Figure 5. The detailed hierarchical agglomerative clustering process is shown in Algorithm \ref{alg_hiercluster}.
\setcounter{algorithm}{1}
\begin{algorithm}[tb]
	\caption{HierCluster}
	\label{alg_hiercluster}
	\textbf{Input}: Dataset $\mathcal{D}_{\text{sample}}$ and distance matrix $\mathbf{M}$ \\
	\textbf{Output}: Clusters $\{\mathbf{S}_1, \mathbf{S}_2, \ldots, \mathbf{S}_K\}$
	\begin{algorithmic}[1] 
		\State Initialize each $x_i \in \mathcal{D}_{\text{sample}}$ as a cluster $\mathbf{S}_i = \{x_i\}$  
		\State $\mathbf{S} \gets \{\mathbf{S}_1, \mathbf{S}_2, \ldots, \mathbf{S}_n\}$ \Comment{Set of clusters}
		\While{True}
		\State $minDist \gets \infty$
		\State $toMerge \gets (null, null)$ 
		\For{each cluster pair $(\mathbf{S}_i, \mathbf{S}_j)$ in $\mathbf{S}$}  
		\State $dist \gets$ Average linkage distance between $\mathbf{S}_i$ and $\mathbf{S}_j$ using $\mathbf{M}$
		\If{$dist < minDist$}
		\State $minDist \gets dist$
		\State $toMerge \gets (\mathbf{S}_i, \mathbf{S}_j)$  
		\EndIf
		\EndFor
		\If{$minDist > t$ or $|\mathbf{S}| == 1$} 
		\State \textbf{break} \Comment{Distance threshold $t$}  
		\EndIf
		\State $\mathbf{S}_{new} \gets$ Merge clusters in $toMerge$
		\State Update $\mathbf{S}$ by removing merged clusters and adding $\mathbf{S}_{new}$
		\EndWhile  
		\State \Return $\{\mathbf{S}_1, \mathbf{S}_2, \ldots, \mathbf{S}_K\}$
	\end{algorithmic}
\end{algorithm}

\begin{figure*}
	\centering
	\includegraphics[width=0.90\linewidth]{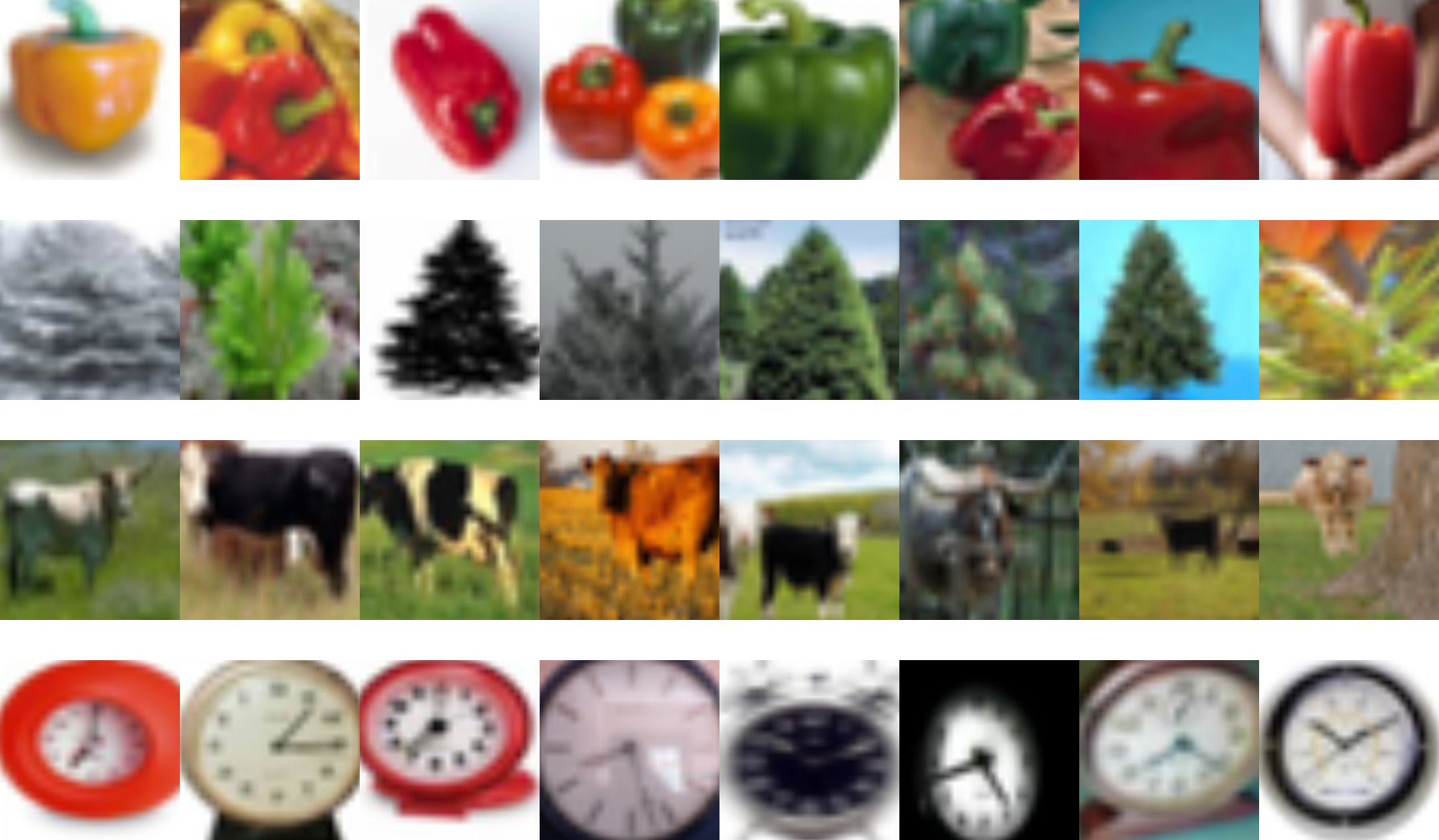}
	\caption{Visualization of HierCluster clustering results. Each row represents a distinct cluster, showcasing images that are grouped based on their diversity.}
	\label{fig:lme-outputcollage}
\end{figure*}
\begin{figure*}
	\centering
	\includegraphics[width=0.90\linewidth]{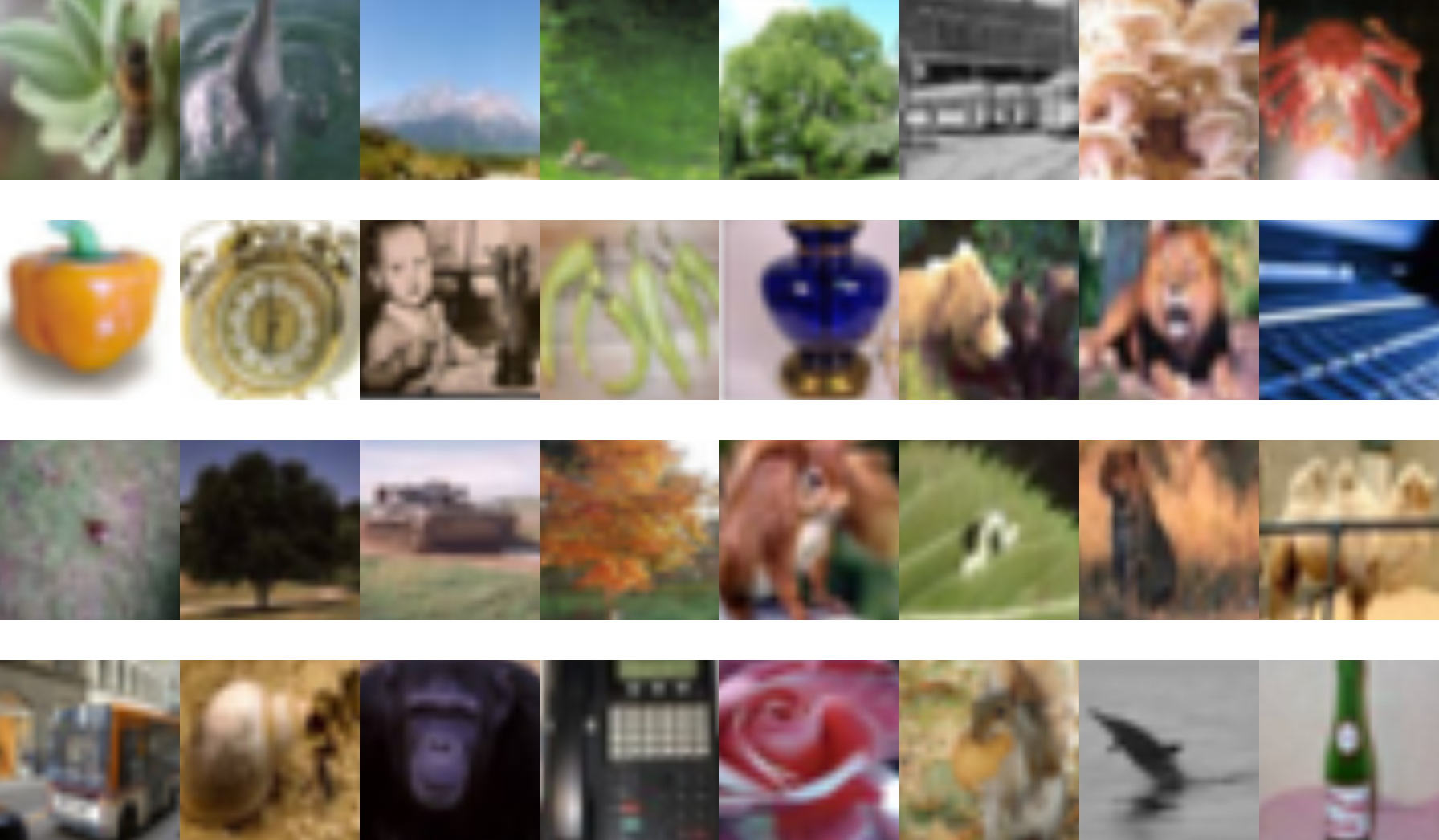}
	\caption{Visualization of random partitioning results. Each row displays images from a randomly assigned group, illustrating the contrast in similarity compared to the HierCluster clustering results.}
	\label{fig:rnd-outputcollage}
\end{figure*}

\subsection{HierCluster Performance}
Our implementation of HierCluster demonstrates its efficiency. In our experimental setup, clustering a sample of 1000 data points from the CIFAR100 dataset takes less than $3$ seconds, approximately 0.1\% of the total time required for 50 epochs of model training. This high efficiency makes the clustering time negligible in the overall process.

The effectiveness of HierCluster is visually presented in Figure \ref{fig:lme-outputcollage}, and for comparison, the results of random partitioning in Figure \ref{fig:rnd-outputcollage}. Figure \ref{fig:lme-outputcollage} illustrates the effectiveness of HierCluster in grouping data points based on their diversity, showcasing the homogeneity within clusters and the distinct separation between them. This clustering approach enables LMEraser to learn more nuanced features within each cluster, significantly enhancing the accuracy of image classification tasks. In contrast, Figure \ref{fig:rnd-outputcollage} demonstrates the results of random partitioning, which lacks such homogeneity and clear separation. This contrast underscores the superiority of the HierCluster approach in effectively grouping similar data points, thereby enabling LMEraser to learn and utilize the intricate diversity of each cluster, improving the accuracy of image classification tasks.

\section{Comprehensive Unlearning Evaluation Metrics}
\subsection{Unlearning Metrics}

Assessing machine unlearning solutions requires a multi-dimensional evaluation across several key metrics. Each metric targets a specific aspect of the unlearning process. 
\begin{itemize} 
	\item \textbf{Data Removal Effectiveness}: This metric evaluates the degree to which the model removes the influence of targeted data points after unlearning. An ideal unlearning method should ensure the model's performance matches what it would be if the removed data had never been used for training.
	\item \textbf{Performance Preservation}: This metric evaluates the model's maintained accuracy on the remaining training data after unlearning. It is important to verify that removing certain data points does not significantly degrade the model's overall performance.
	\item \textbf{Unlearning Time Efficiency}:  This metric evaluates the time it takes to unlearn the data. Efficient unlearning methods should ideally require less time than retraining the model from scratch. 
	\item \textbf{Resource Consumption}: This metric evaluates the computational and storage resources demanded by the unlearning process. An optimal unlearning approach is characterized by its efficiency in minimizing the use of computational power and storage, thereby enhancing the overall feasibility and sustainability of the unlearning mechanism in diverse operational environments.
	\item \textbf{Scalability}: This metric evaluates the unlearning method's adaptability to varying dataset sizes, model complexities, and the ability to manage multiple unlearning requests concurrently. Scalability is critical for ensuring the method's applicability to large-scale and complex models.
\end{itemize}

These metrics collectively provide a comprehensive framework for assessing machine unlearning solutions across key practical dimensions. 

\subsection{Assessing LMEraser}  
LMEraser stands out in its efficient performance across various critical metrics.

\begin{itemize}
	\item \textbf{Data Removal Effectiveness}: LMEraser achieves exact unlearning by isolating the influence of data points through the adaptive tuning mechanism, removing the target private data points from clusters, and re-optimizing the affected prompts and heads. 
	\item \textbf{Performance Preservation}: LMEraser ensures minimal performance degradation. As shown in Figure 5, when unlearning 50\% of the CIFAR100 dataset, test accuracy was slightly decreased, from 0.7275 to 0.6953. Across tests, it retained over 95\% of remaining data accuracy after unlearning half the private dataset, showcasing its robustness.
	\item \textbf{Unlearning Time Efficiency}: The time efficiency of LMEraser is highlighted by its rapid unlearning process, as evidenced in Table 2. It removes individual data points in tens of seconds, contrasting the days required for full retraining and showcasing its rapid adaptation capabilities. 
	\item \textbf{Resource Consumption}:   LMEraser's approach drastically cuts down computational demands by requiring only the retraining of prompt parameters rather than the entire model. Figure 6  shows that LMEraser retrains approximately 200K parameters compared to the 85M needed for naive retraining, reducing computational costs by approximately 425-fold.
	\item \textbf{Scalability}: As demonstrated in Figures 3 and 4, LMEraser effectively scales to large datasets and complex models, handling up to 60,000 images and models with billions of parameters. This scalability ensures its applicability and effectiveness in large-scale implementations. 
\end{itemize}

LMEraser has been comprehensively evaluated on image classification tasks, showcasing its efficient and exact unlearning in practical scenarios. The core of its effectiveness lies in an adaptive prompt tuning mechanism. This mechanism ensures minimal degradation in the accuracy of the remaining data while significantly reducing computational demands. Moreover, LMEraser exhibits remarkable adaptability, seamlessly accommodating large datasets and complex model architectures.

\bibliographystyle{IEEEtran}
\bibliography{ijcai24}

\vfill

\end{document}